\documentclass[11pt,notitlepage,twoside,a4paper]{article}

\usepackage{hyperref}
\usepackage{listings}
\usepackage{color}
\usepackage{xspace}
\usepackage{graphicx}

\newcommand{\ignore}[1]{}

\newcommand{\set}[1]{\ensuremath{\mathbf{#1}}\xspace}
\newcommand{\code}[1]{\texttt{#1}\xspace}
\newcommand{\seq}[2]{\ensuremath{\{#1,\allowbreak\dots,#2\}}\xspace}                                                        

\begin{document}

\author{Tommaso Urli \thanks{\emph{Scheduling and Time-Tabling Group}, DIEGM - University of Udine, Via delle Scienze 206, 33100 -- Udine (UD), Italy}\\\texttt{\href{mailto:tommaso.urli@uniud.it}{tommaso.urli@uniud.it}}}
\title{\textsc{Technical Report}\\Balancing bike sharing systems (BBSS): instance generation from the CitiBike NYC data}

\maketitle

\tableofcontents

\section{Introduction}

    Bike sharing systems are a very popular means to provide bikes to citizens in a simple and cheap way. The idea is to install bike stations at various points in the city, from which a registered user can easily loan a bike by removing it from a specialized rack. After the ride, the user may return the bike at any station (if there is a free rack). Services of this kind are mainly public or semi-public, often aimed at increasing the attractiveness of non-motorized means of transportation, and are usually free, or almost free, of charge for the users. 
    
    Depending on their location, bike stations have specific patterns regarding when they are empty or full. For instance, in cities where most jobs are located near the city centre, the commuters cause certain peaks in the morning: the central bike stations are filled, while the stations in the outskirts are emptied. Furthermore, stations located on top of a hill are more likely to be empty, since users are less keen on cycling uphill to return the bike, and often leave their bike at a more reachable station.
    These issues result in substantial user dissatisfaction which may eventually cause the users to abandon the service. This is why nowadays most bike sharing system providers take measures to \emph{rebalance} them. Balancing a bike sharing system is typically done by employing a fleet of trucks that move bikes overnight between unbalanced stations. More specifically, each truck starts from a depot and travels from station to station in a tour, executing loading instructions (adding or removing bikes) at each stop. After servicing the last station, each truck returns to the depot.
    
    Over the last few years, balancing bike sharing systems (BBSS) has become increasingly studied in optimization \cite{BeCDLM11, CoMR12, RaTF12, ChMWC12, ScHH13, DiRU13a, DiRU13b, RaPHR13}. As such, generating meaningful instance to serve as a benchmark for the proposed approaches is an important task.
    In this technical report we describe the procedure we used to generate BBSS problem instances from data of the \href{http://www.citibikenyc.com}{CitiBike NYC} bike sharing system.

\section{Instance format}

    We employ the instance format\footnote{We refer the reader to \href{https://www.ads.tuwien.ac.at/w/Research/Problem_Instances}{https://www.ads.tuwien.ac.at/w/Research/Problem\_Instances} for a complete description of the format and all its variants, e.g., the dynamic formulation.} defined and popularized by the \href{https://www.ads.tuwien.ac.at/w/Arbeitsbereich_f%C3%BCr_Algorithmen_und_Datenstrukturen}{ADS} group at the Technische Universit\"at Wien (TU Wien) and the \href{http://www.ait.ac.at/departments/mobility}{mobility department} at Austrian Institute of Technology (AIT). Note, however, that our scope is limited to the static variant of the problem, as such we only consider a valid subset of the instance format.
    
    The format for the static BBSS specifies, for each station $s \in \set{S}$
    
    \begin{itemize}
        \itemsep -0.2em
        \item the \textbf{current number of bikes} $b_s$,
        \item the \textbf{target number of bikes} $\hat{b}_s$,
        \item the \textbf{distance from the depot} $d_{s,d}$, and
        \item the \textbf{distance from each other station $k$} $d_{s,k}$.
    \end{itemize}
    
    Note that this format only describes a state of the bike sharing system, therefore, in a sense, it describes a \emph{family} of instances. Specific instances can be generated by specifying
    
    \begin{itemize}
        \itemsep -0.2em
        \item the \textbf{stations capacities} $C_s$,
        \item the \textbf{vehicles capacities} $c_v, v \in \set{V}$,
        \item the \textbf{number of vehicles from the depot} $V$, and
        \item the \textbf{vehicles time budget} $\hat{t}_v$.
    \end{itemize}
            
\section{Data collection}

    The first step of instance generation, is gathering a sufficient amount of usage data about a bike sharing system. Our choice system of choice is CitiBike NYC\footnote{CitiBike NYC:  \href{http://www.citibikenyc.com}{http://www.citibikenyc.com}}, since they provide full access, through a web service, to the state of the network in JSON format at any time\footnote{Web service URL: \href{http://citibikenyc.com/stations/json}{http://citibikenyc.com/stations/json}}. This is an example of output from the web service
    
    \begin{small}
        \begin{verbatim}
    {
        "executionTime": "2013-11-04 12:09:01 AM", 
        "stationBeanList": [
            {
                "availableDocks": 21, 
                "totalDocks": 39, 
                "longitude": -73.99392888, 
                "testStation": false, 
                "stAddress1": "W 52 St & 11 Ave", 
                "stationName": "W 52 St & 11 Ave", 
                "landMark": "", 
                "latitude": 40.76727216, 
                "statusKey": 1, 
                "availableBikes": 17, 
                ...
                "id": 72
            }, 
            ...
        ]
    }
        \end{verbatim}
    \end{small}
    From the output, it can be noted that three pieces of information about capacity are reported: the \code{availableDocks}, \code{totalDocks}, and the \code{availableBikes}. One interesting thing is that

    \begin{equation}
        totalDocks \not= availableDocks + availableBikes
    \end{equation}
    i.e., there is a displacement of one bike, which is, however, constant through the stations. Moreover, a field \code{statusKey} encodes the status of the station, e.g., operational or non-operational.
        
    We have stored a snapshot of the network every $10$ minutes for 6 months (since May 2013 to November 2013). A snapshot, among other information, contains the current and the maximum number of bikes of each station, and its address. This data was necessary in order to provide realistic $b_s$ and $\hat{b}_s$ for every station. Moreover, we employed the address data to query the Google Maps API\footnote{Google Maps API: \href{https://developers.google.com/maps}{https://developers.google.com/maps}} about the distance, both in minutes and meters, between each pair of stations $s, k \in \set{S}$. Note that this also includes the distances from a depot, as we consider one of the central stations as a depot.

\section{Data processing}
    
\begin{figure}[!ht]
        \includegraphics[width=\textwidth]{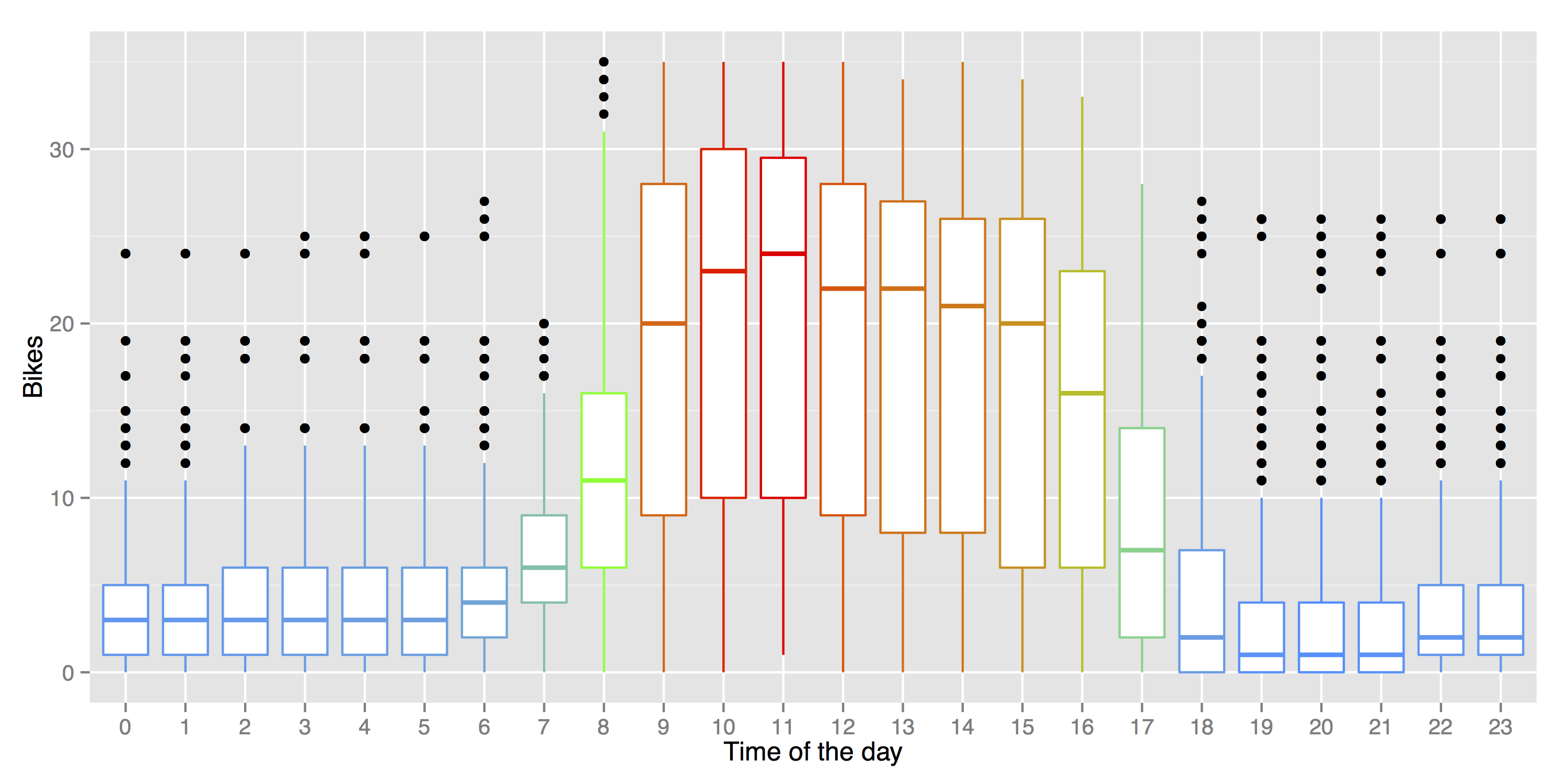}
        \caption{Example of source station \label{fig:source}}
    \end{figure}

\begin{figure}[!ht]
        \includegraphics[width=\textwidth]{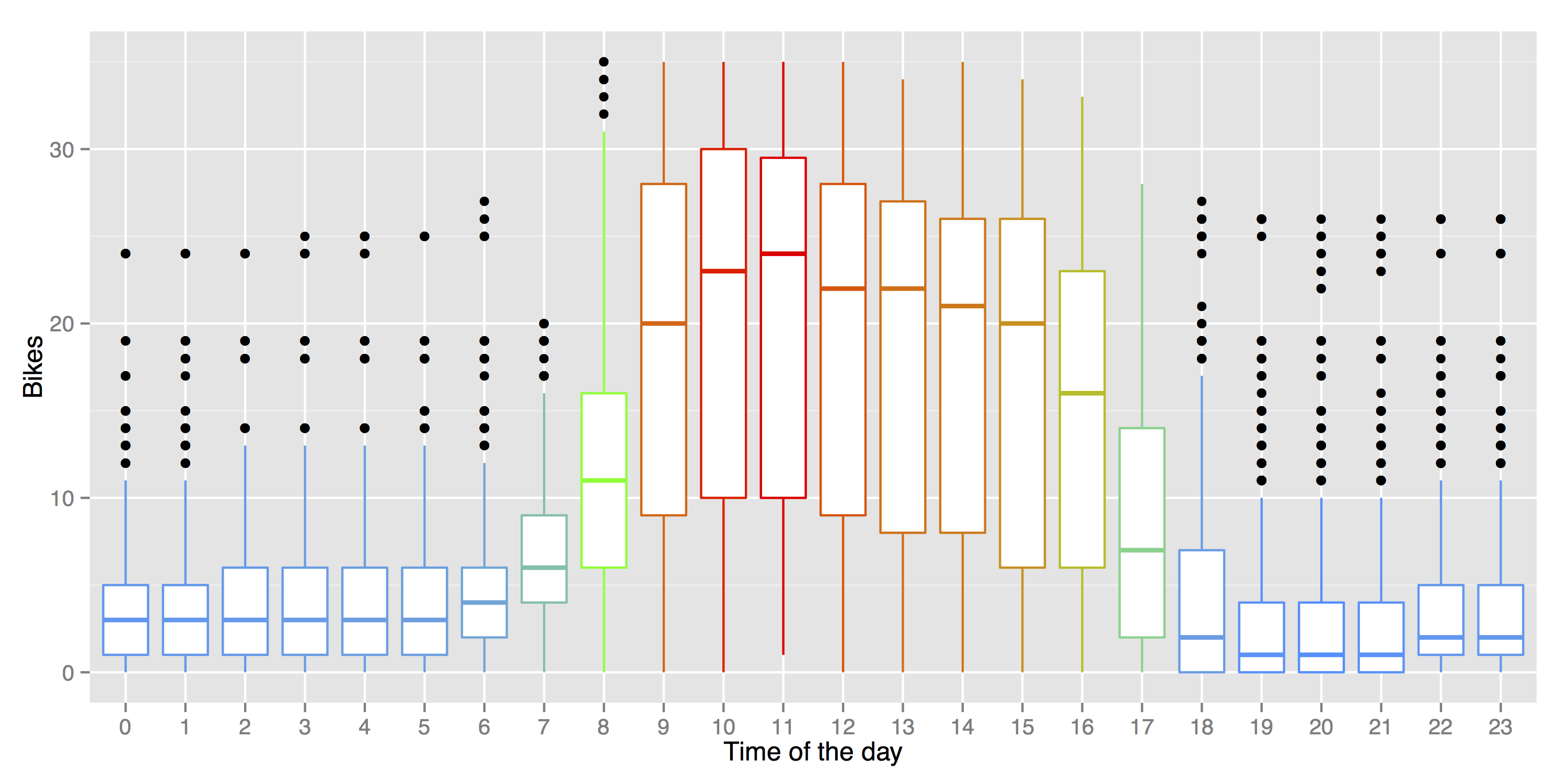}
        \caption{Example of sink station \label{fig:sink}}
    \end{figure}

    We considered the $\approx 25'000$ snapshots, and, for each station, we computed the distributions of stored bikes at every hour of the day. The week-ends were not considered because the bike usage is much noisier than during working days. From the analysis it is clear that, at certain times of the day, there are stations acting as \emph{sources} (see Figure~\ref{fig:source}) and others acting as \emph{sinks} (see Figure~\ref{fig:sink}). The following box plots, that show the distributions of bikes on a single station throughout the day show this behavior.

\begin{figure}[!ht]
        \includegraphics[width=\textwidth]{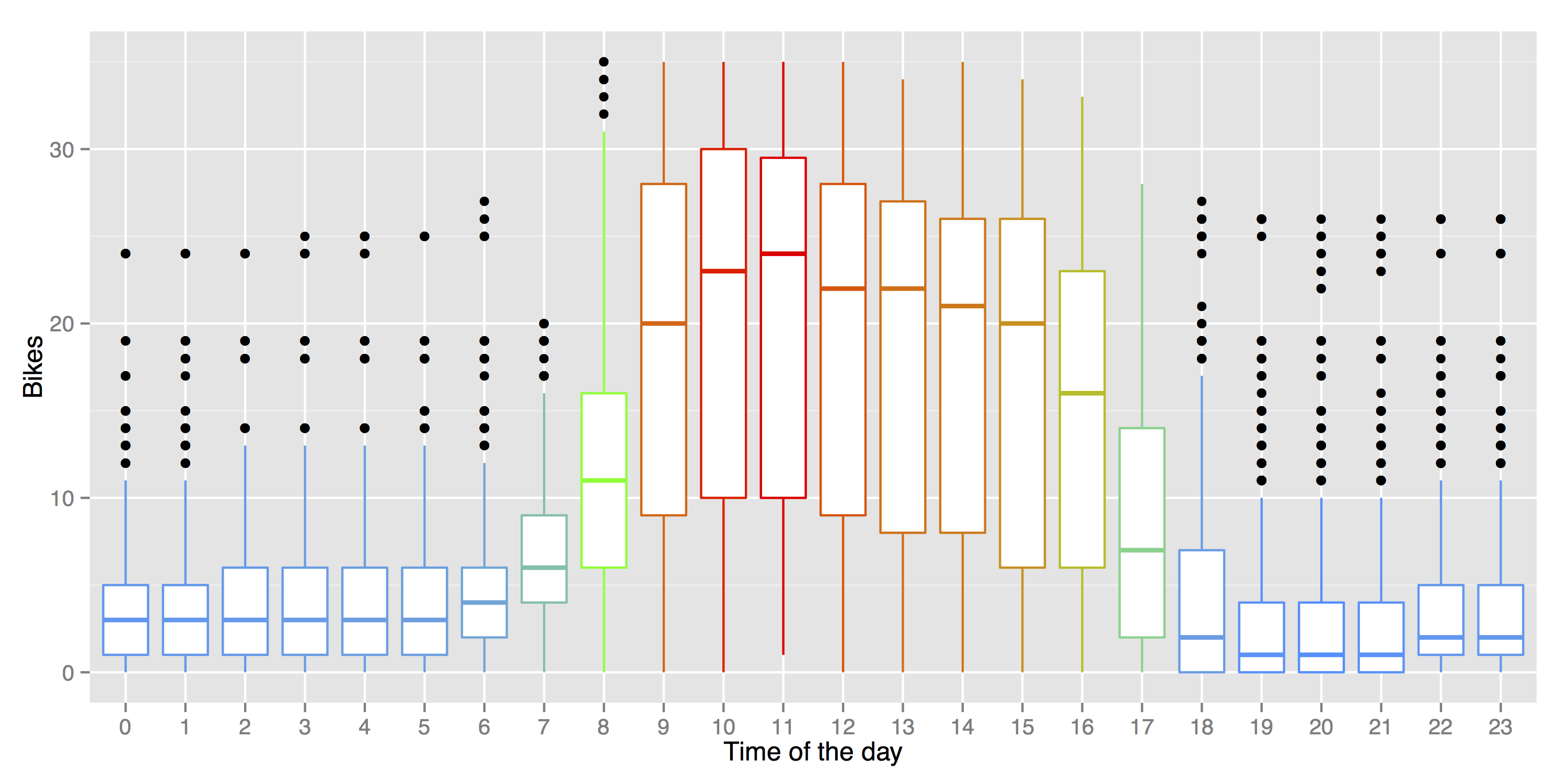}
        \caption{Rebalancing step around 04:00 AM \label{fig:step}}
    \end{figure}

    \noindent
    From some of the distributions, it was also clear that there was some artificial rebalancing happening overnight between $00:00$ and $06:00$ AM (mildly visible in Figure~\ref{fig:step} at $04:00$ AM).

    For each station $s \in \set{S}$, we first computed the $1^{st}$ and $3^{rd}$ quartiles for all the hours of the day. Then we found the minimum first quartile and the maximum third quartile across the day, which we denote, respectively, by $min_s$ and $max_s$. Ideally, these are values which we would like to be as far as possible, respectively from $0$ (empty station) and from $C_s$ (full station). We thus computed a displacement value for each station $s$, as
    
    \begin{equation}
        disp_s = \left\lfloor C_s - (C_s - (max_s-min_s)) / 2 \right\rfloor - max_s
    \end{equation}
    ideally, displacing the distributions by $disp_s$ brings $min_s$ and $max_s$ as far as possible from $0$ and $C_s$, so that the probability of finding an empty or a full station is minimized.
    
\section{Instance generation}

    Once the displacement of each station is known, generating an instance from a snapshot is rather easy. But there are some aspects which one should take into account.
    
    \paragraph{Selection of the snapshot.} In order for the generated instances to be realistic, one should consider when the rebalancing is likely to be done. A good guess for this is that the rebalancing happens overnight. This is also supported by the \emph{rebalancing step} which is visible on some stations (e.g., Figure~\ref{fig:step}). For our instances, we have chosen midnight as the expected time for the \emph{start} of the rebalancing, thus we have used the $30$ midnight snapshots from September 2013 as starting points. For each station $s \in \set{S}$, the initial number of bikes $b_s$ is thus the actual number of bikes in the station at midnight.
    
    \paragraph{Selection of the depot.} The information released by CitiBike NYC does not contain any data about depots. Of course, this station must be excluded from the choice of the other stations to include in the instance. We selected the station with CitiBike ID $294$ as it was quite centra with respect to all the other stations.
    
    \paragraph{Selection of stations.} Our generator accepts a \code{size} parameter that controls the number of stations that are included in the instance. The stations are then partitioned in \emph{sinks} and \emph{sources} and a random station from each set is added uniformly at randomly to the instance. This strategy tries to balance sinks and sources, so that the objective function range is broader (and the generated instances are more interesting).
    Note that, because of the randomness in the generation process, the generated instance use, in principle, a different set of stations. However, since the number of stations in the CitiBike NYC sistem is limited ($\approx 330$) the larger instances are more likely to include similar sets of stations.

\section{Generator and instances}

    We have built an instance generator for BBSS based on the ideas described in the previous section. The generator is available under the permissive MIT license at the address \href{https://bitbucket.org/tunnuz/citibike-nyc-generator}{https://bitbucket.org/tunnuz/citibike-nyc-generator}. Moreover, we have generated $180$ instances of increasing size $\in \seq{40, 80}{240}$, which are publicly available as well, under the same license at the address \href{https://bitbucket.org/tunnuz/citibike-nyc-sept-13}{https://bitbucket.org/tunnuz/citibike-nyc-sept-13}. Note that, unlike the instances available from the ADS and AIT websites, the one generated by our software do not already consider the bicycle loading / unloading times inside the traveling times. The distances are thus real distances, and the loading and unloading times must be added to the cost function. This allows to implement various policies, e.g., fixed loading / unloading times, or loading / unloading time dependent on the number of transferred bicycles. 

\bibliographystyle{plain}
\bibliography{paper}

\end{document}